\pdfoutput=1

\documentclass[11pt]{article}

\usepackage[]{ACL2023}

\usepackage{times}
\usepackage{latexsym}

\usepackage{amsmath,amsfonts}
\usepackage{algorithmic}
\usepackage{algorithm}
\usepackage{array}
\usepackage[caption=false,font=normalsize,labelfont=sf,textfont=sf]{subfig}
\usepackage{textcomp}
\usepackage{stfloats}
\usepackage{url}
\usepackage{verbatim}
\usepackage{graphicx}
\usepackage{cite}
\usepackage{multirow}

\usepackage[T1]{fontenc}

\usepackage[utf8]{inputenc}

\usepackage{microtype}

\usepackage{inconsolata}

\title{Disambiguated Lexically Constrained Neural Machine Translation}

\author{
	{Jinpeng Zhang\textsuperscript{1}, Nini Xiao\textsuperscript{1}, Ke Wang\textsuperscript{2}, Chuanqi Dong\textsuperscript{1}, Xiangyu Duan\textsuperscript{1}\thanks{  $\quad $ Corresponding Author. } } \\
	{\textbf{Yuqi Zhang\textsuperscript{2}, Min Zhang\textsuperscript{1} } }
	\vspace{2.0mm}\\
	\fontsize{12}{10}\selectfont
	\,\textsuperscript{\rm 1}  Institute of Aritificial Intelligence, School of Computer Science and Technology, \\
	\fontsize{12}{10}\selectfont Soochow University  \\
            \fontsize{12}{10}\selectfont  \textsuperscript{\rm 2} Alibaba DAMO Academy \\
            \fontsize{10}{10}\selectfont jpzhang1@stu.suda.edu.cn; xiangyuduan@suda.edu.cn}

\begin{document}
\maketitle
\begin{abstract}
Lexically constrained neural machine translation (LCNMT), which controls the translation generation with pre-specified constraints, is important in many practical applications. Current approaches to LCNMT typically assume that the pre-specified lexical constraints are contextually appropriate. This assumption limits their application to real-world scenarios where a source lexicon may have multiple target constraints, and disambiguation is needed to select the most suitable one. In this paper, we propose disambiguated LCNMT (D-LCNMT) to solve the problem. D-LCNMT is a robust and effective two-stage framework that disambiguates the constraints based on contexts at first, then integrates the disambiguated constraints into LCNMT. Experimental results show that our approach outperforms strong baselines including existing data augmentation based approaches on benchmark datasets, and comprehensive experiments in scenarios where a source lexicon corresponds to multiple target constraints demonstrate the constraint disambiguation superiority of our approach.
\end{abstract}

\section{Introduction}

Lexically constrained neural machine translation (LCNMT) is a task that guarantees the inclusion of specific lexicons in the translation, which is of great importance in many applications such as interactive translation with user-given lexicon constraints \citep{koehn2009process}, domain adaptation with pre-specified terminology constraints \citep{hasler-etal-2018-neural}. Accurate lexicon translation plays a key role in improving translation quality. However, in real world applications, a source lexicon often has multiple translation constraints, which are provided by a specific database and represent different but core concepts. It is essential for a translation model to select the most contextually appropriate constraint and force it to appear in the translation, but such constraint disambiguation process is largely ignored in previous LCNMT researches. They just use the aligned target lexicons appeared in the translation reference of a given source sentence as the constraints and bypass the constraint ambiguity problem \citep{dinu-etal-2019-training,song-etal-2019-code,wang2022template,wang-etal-2022-integrating}. In this paper, we propose disambiguated LCNMT (D-LCNMT) to solve the constraint ambiguity problem when facing a source sentence, and investigate how to integrate the disambiguated constraints into NMT.

\begin{figure}[!t]
\centering
\includegraphics[width=3.1in]{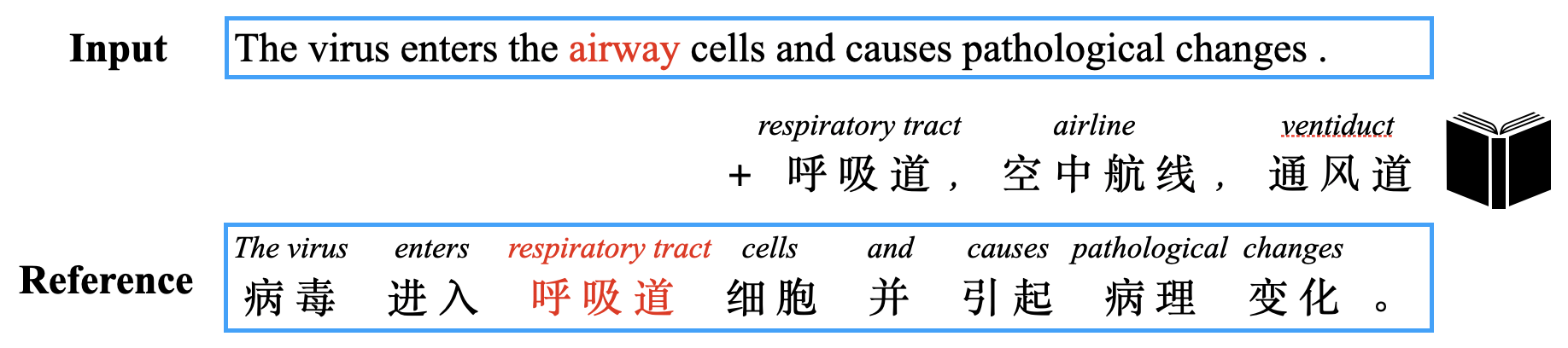}
\caption{An example of the constraint ambiguity problem in English-to-Chinese translation. Given a lexical constraint inventory, the lexicon \emph{airway} has three possible translations as the ambiguous constraints: \emph{respiratory tract, airline}, and \emph{ventiduct}, among which \emph{respiratory tract} is the context appropriate one for the input sentence.}
\label{fig_example}
\end{figure}

\begin{table}[!t]
\caption{The frequency of the constraint ambiguity problem in the validation sets of German-to-English(De-En) and English-to-Chinese(En-Zh) translation tasks. \label{tab_const_num}}
\small
\centering
\begin{tabular}{c|| c|c}
\hline
  & Ambiguous Constraints & Total Constraints \\
\hline
De-En & 1146 & 2243 \\

En-Zh & 566 & 743 \\
\hline
\end{tabular}
\end{table}

Figure \ref{fig_example} presents an example of the constraint ambiguity problem. Table \ref{tab_const_num} presents the frequency of the problem in the validation sets, showing that the ambiguous constraints account for more than half of the total constraints. Despite the severity of the problem, it is overlooked by most LCNMT researches which only use gold constraints. The problem is brought into the spotlight only at recent WMT2021 shared task on machine translation using terminologies, where a source terminology has averagely 2.22 possible translation constraints. Major works in this task apply data augmentation approach, which builds synthetic corpora containing ambiguous constraints via code-switching, and train the NMT models to select the most contextually appropriate constraint implicitly \citep{wang-etal-2021-termmind, ailem-etal-2021-lingua}. 

Instead, our D-LCNMT adopts an explicit two-stage framework that performs constraint disambiguation and integration into NMT sequentially, and outperforms the above data augmentation approach on benchmark datasets. In particular, at the first stage, we build a constraint disambiguation network based on contrastive learning so that the correct constraint is selected given the source lexicon and its context in the given source sentence. At the second stage, we integrate the most appropriate constraint obtained in the first stage into NMT with the help of current lexically constrained approaches \citep{wang2022template,wang-etal-2022-integrating}. Experiments on disambiguated lexically constrained translation tasks in German-to-English and English-to-Chinese show that our approach significantly outperforms strong baselines including the data augmentation approach. For lexicons that have multiple possible constraints, our approach achieves  state-of-the-art  accuracy of constraint disambiguation, especially ranks the first in the leaderboard of WMT2021 shared task on machine translation using terminologies. Overall, our contributions are three-fold:

\begin{enumerate}
\item{We propose D-LCNMT which is a robust and effective two-stage framework that disambiguates the constraints at first, then integrate the constraints into LCNMT.}
\item{We propose a continuous encoding space with contrastive learning for constraint disambiguation, which is a problem overlooked by major LCNMT researches which use gold constraints.}
\item{Through extensive evaluation and comparison to other approaches, we achieve the best constraint disambiguation accuracy, and maintain or achieve higher sentence level translation quality.}
\end{enumerate}

\section{Related Work}

We introduce LCNMT at first, then introduce the related constraint disambiguation researches. 

\subsection{LCNMT}

LCNMT controls the translation output of an NMT model to satisfy some pre-specified lexical constraints. The lexical constraints are usually provided by users or deposit dictionaries covering wide range of topics and domains, showing great values in practical applications.

One line of LCNMT studies focuses on designing constrained decoding algorithm \citep{hasler-etal-2018-neural}. For example, \citet{hokamp-liu-2017-lexically} firstly proposed grid beam search (GBS), which added an additional dimension of the number of constrained lexicons at each decoding step. \citet{post-vilar-2018-fast} proposed a dynamically beam allocating (DBA) strategy for constrained decoding, which fixed the beam size and made it unaffected by the number of constrained lexicons. Then, \citet{hu-etal-2019-improved} extended it into vectorized dynamic beam allocation (VDBA) that supports batched decoding. Although these constrained beam search methods have high control over the target constraints, they significantly slow down the decoding speed and tend to reduce the fluency of translation \citep{hasler-etal-2018-neural}.

Another line of studies addresses the problem by augmenting the training data with placeholders or additional translation constraints. \citet{crego2016systran} proposed to replace entities with placeholders, which remained in the system output. They are placed back through post-processing. \citet{song-etal-2019-code} replaced the source lexicons with the corresponding target constraints, and \citet{dinu-etal-2019-training} appended the target constraints right after the corresponding source lexicons. During inference, the target constraints are imposed on the source sentences similarly. The main disadvantage of these methods is that they do not guarantee the appearance of the target constraints in some cases \citep{chen2021lexically}.

Different from the above decoding and synthetic data approaches, models of constrained neural networks were also explored. \citet{song2020alignment} trained an alignment-enhanced NMT model and conducted alignment-based constrained decoding, but they required alignment labels from external aligners with noisy alignments. \citet{susanto-etal-2020-lexically} proposed to invoke constraints using a non-autoregressive approach, while the constraints must be in the same order to that in reference. \citet{wang-etal-2022-integrating} vectorized source and target constraints into continuous keys and values and integrated them into NMT. Recently, \citet{wang2022template} proposed a template-based constrained translation framework to disentangle the generation of constraints and free tokens, and achieved high translation quality and constraint match accuracy with inference speed unchanged.

\begin{figure*}[!t]
\centering
\includegraphics[width=6.5in]{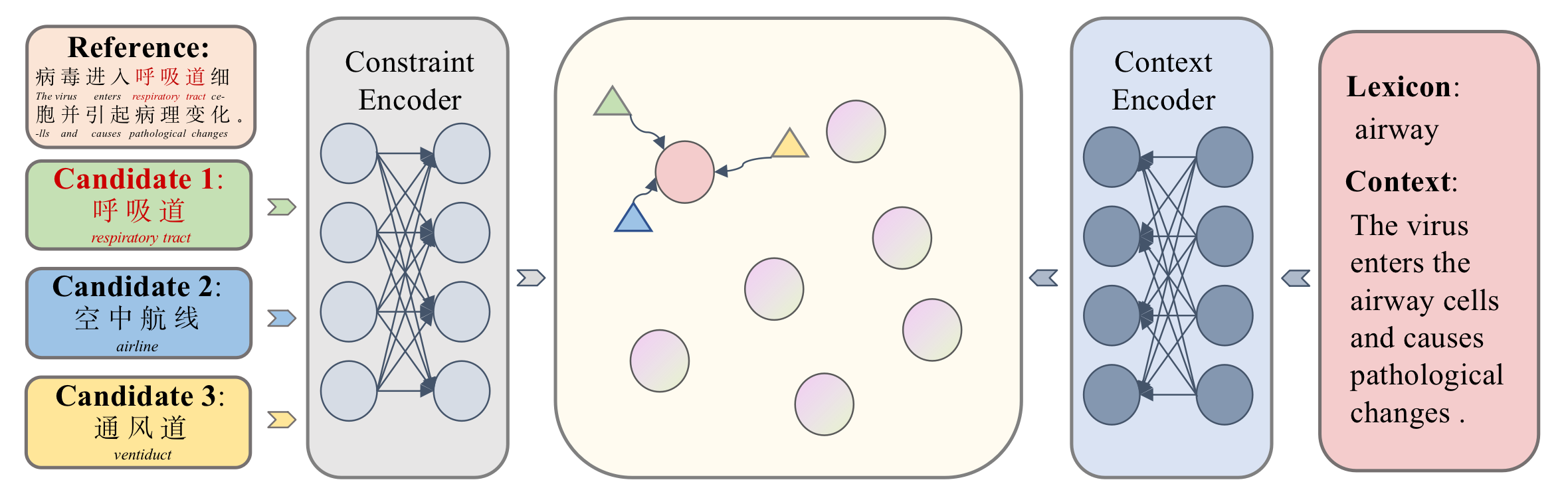}
\caption{The constraint disambiguation neural network. Given the source lexicon \emph{airway} and its context shown in the right, the framework selects the correct constraint \emph{respiratory tract} from all three candidate constraints by building the common representation space for the source and target sides as shown in the middle. }
\label{disambugation}
\end{figure*}

\subsection{Constraint Disambiguation}

The above studies on LCNMT assume that the pre-specified lexical constraints are gold ones. For a source sentence, the constraints are simulated by being directly extracted from the target sentence. Such simulation is not practical when a source lexicon has multiple possible translations as constraints, and the target sentence is not known when translating an input source sentence. This ambiguous constraint problem for LCNMT is noticed by researchers at the WMT2021 shared task on machine translation using terminologies, where certain terminologies have multiple possible translations as the ambiguous constraints. \citet{ailem-etal-2021-lingua} solve the problem by selecting terminology translations at random and insert them as constraints in the source sentence. \citet{wang-etal-2021-termmind} propose to augment source sentence with all possible terminology translations, which is different from \citet{ailem-etal-2021-lingua} who kept only one. These data augmentation methods do not explicitly disambiguate the constraints. They just train an NMT model to generate correct sentence level translations given the augmented source sentence. Unlike previous works, we propose an explicit constraint disambiguation module to select the most contextually appropriate constraint. 

\section{D-LCNMT}

We propose D-LCNMT to solve the ambiguous constraint problem for LCNMT through a two-stage framework. At Stage 1, we introduce a contrastive learning based constraint disambiguation neural network. At Stage 2, we integrate the disambiguated constraints into current competitive LCNMT models \citep{wang2022template,wang-etal-2022-integrating}. 

\begin{figure*}[!t]
\centering
\includegraphics[width=6.5in]{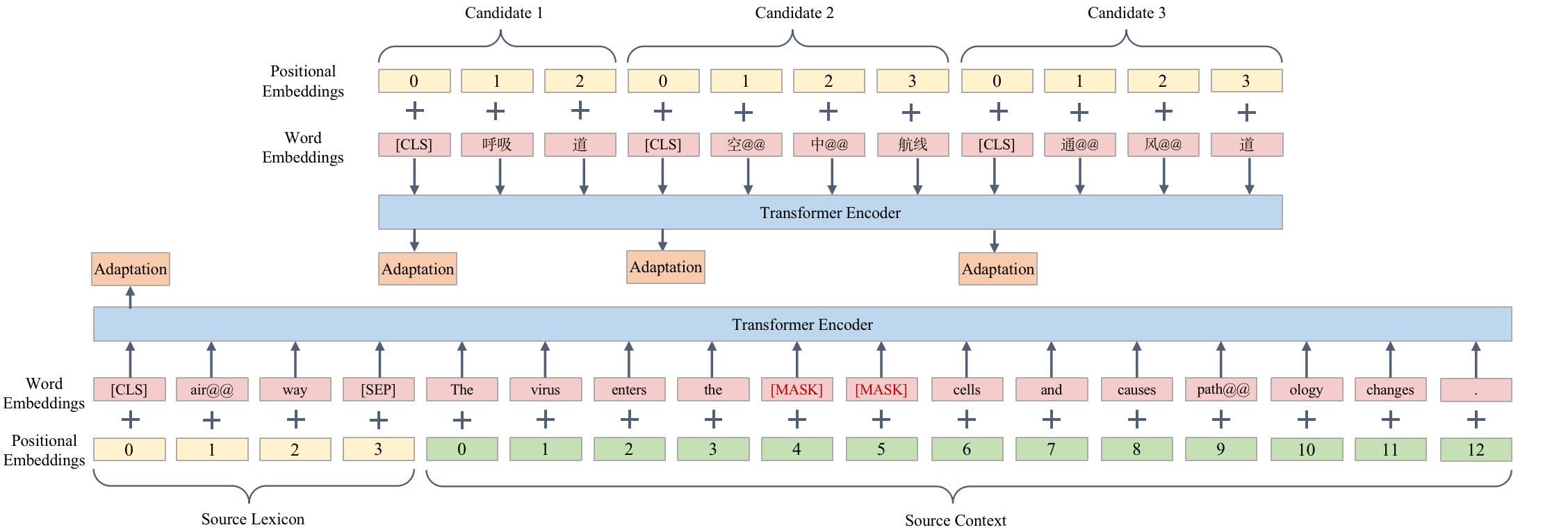}
\caption{The structure of the context encoder and the constraint encoder.}
\label{fig_context}
\end{figure*}

\subsection{Stage 1: Constraint Disambiguation}

In a lexical constraint inventory, which is provided by either users or dictionaries, a source lexicon may have multiple possible translations. Let ${\bf s}^{}$ denotes a source lexicon, its ambiguous translations are ${\bf m}_{{\bf }^{}}^{(1)},...,{\bf m}_{{\bf }^{}}^{(K)}$. The constraint disambiguation is needed to select one appropriate translation given ${\bf s}^{}$ and its source context ${\bf C}_{{\bf s}^{{}}}$ in the input source sentence. 

The constraint disambiguation neural network is shown in Fig.~\ref{disambugation}. The main goal is to encode the source lexicons with contexts and the corresponding target side candidate constraints into the same representation space, so that the source lexicons and their correct constraints are closest neighbors in the space. Briefly, the network consists of a context encoder and a constraint encoder. In the source side, the context encoder captures the semantic information of the lexicons and their contexts at the same time. In the target side, the constraint encoder considers all possible candidate constraints for a source lexicon and encode each variable-length candidate into a single representation.

\paragraph{Context Encoder and Constraint Encoder} Both encoders are independent of each other. Each of them consists of two transformer encoder layers stacked with one adaptation layer. For either the source lexicon or its translation constraint, we add a special token [CLS] in front of it. The hidden state of [CLS] outputted by the encoder is used as its representation.

For a considered source lexicon, we concatenate it with the source sentence by adding a special token [SEP] and feed the concatenation to the context encoder to obtain the representation of the source lexicon. The structure is shown in Fig.~\ref{fig_context}. Notably, the source lexicon is masked in the source sentence to let the encoder better encode the context of the lexicon. The positions of the lexicon and the sentence are independently countered. For each translation constraint, we directly feed it to the constraint encoder, and get the hidden state of [CLS] as its representation.

In each encoder, the adaptation layer is stacked over the transformer layers for further optimizing the hidden state of [CLS]. The adaptation layer consists of two linear transformations and a tanh activation in between \citep{wang-etal-2022-integrating}. Let ${\bf h}_{{\bf s}_{}}\in \mathbb{R}^{d \times 1}$ and ${\bf h}_{{\bf m}_{}^{(k)}}\in \mathbb{R}^{d \times 1}$ be the two hidden states of [CLS] outputted by the transformer layers in the source and target side, respectively. The final outputs of the context encoder and the constraint encoder are defined as:

\begin{equation}
\begin{aligned}
\label{template1}
\begin{array}{rl}
& {\bf e}_{{\bf s}^{}}={\rm \tanh}({\bf h}_{{\bf s}^{}}^{\rm T} {\bf W}_{1}){\bf W}_{2},\\
& {\bf e}_{{\bf m}^{(k)}_{\bf }}={\rm tanh}({\bf h}_{{\bf m}^{(k)}_{\bf }}^{\rm T} {\bf W}_{3}){\bf W}_{4},\\
\end{array} 
\end{aligned}
\end{equation}
where ${\bf W}_{\cdot} \in \mathbb{R}^{d \times d}$ presents the trainable linear transformations.

\paragraph{Contrastive Learning} Contrastive learning can learn effective representation by pulling semantically close neighbors together and pushing apart non-neighbors \citep{gao-etal-2021-simcse, pan-etal-2021-contrastive}. We adopt the contrastive objective to train the disambiguation network. For a given parallel sentence pair, we treat the source lexicon ${\bf s}$ and its translation ${\bf t}^{}$ in the target sentence as positive constraint sample, and treat ${\bf s}$ and its other candidate translations as negative constraint samples. Let ${\bf e}_{{\bf s}^{}}, {\bf e}_{{\bf t}^{}}$ be the representation of ${\bf s}^{}$ and ${\bf t}^{}$, respectively. The training loss for each sentence pair is:  
\begin{equation}
\label{ctrloss2}
L_{\rm ctr} = -\sum_{n=1}^{N}\log \frac{e^{\text {sim}{(
{\bf e}_{{\bf s}^{(n)}},
{\bf e}_{{\bf t}^{(n)}}
)}}}{\sum_{k=1}^{K}e^{\text{sim}{(
{\bf e}_{{\bf s}^{(n)}},
{\bf e}_{{\bf m}^{(k)}_{(n)}}
)}}}.
\end{equation}
where $\rm sim(\cdot)$ denotes the cosine function, $N$ is the number of constraints contained the training parallel sentences. In practice, there are some source lexicons having too many or few candidate translations, which may affect the performance of the contrastive learning. To address this issue, for each of such source lexicons, we randomly select $K$ candidate translations of it derived from the predefined inventory as negative samples. If a source lexicon has less than $K$ candidate translations, we randomly select other translations from the training batch to complement $K$ negative samples. During inference, we calculate the cosine similarity between the source representation and each constraint candidate representation, and select the one with the highest cosine similarity as the disambiguated constraint.

\subsection{Stage 2: Integrating Disambiguated Constraints into LCNMT}

At Stage 2, we choose two recent competitive LCNMT systems, which are originally developed for integrating gold constraints, to integrate our disambiguated constraints. One is VecConstNMT \citep{wang-etal-2022-integrating}, which is based on constraint vectorization and outperforms several strong baselines. However, we found that VecConstNMT failed in copying long constraints integrally due to its word-by-word generation nature. To address this issue, we propose an integrity loss and a decoding strategy to ensure the appearance of the long constraints in translation. The other is template-based LCNMT \citep{wang2022template}, which achieves high translation quality with 100\% success rate of generating the constraints. So we simply feed the disambiguated constraints directly into the template-based LCNMT.

\paragraph{Integration into VecConstNMT} 

VecConstNMT splits the translation probability into two subparts: $P_{\rm model}$ and $P_{\rm plug}$, where $P_{\rm model}$ is the conventional form of Transformer probability, $P_{\rm plug}$ is the probability tailored for the lexical constraints. Suppose a sentence pair $\langle {\bf x}, {\bf y}\rangle$ with $N$ lexical constraints (${\bf s}_1^{N}$, ${\bf t}_1^{N}$) \footnote{During training, we use gold constraints contained in the sentence pair. During testing, we use the disambiguated constraints generated by Stage 1.}:
\begin{equation}
\begin{aligned}
\label{pmodel}
P_{\rm model}(y|{\bf{y}}_{<i}, {\bf{x}}, {\bf s}_1^{N}, &{\bf t}_1^{N}; \bf {\theta}) \\
&=\rm{softmax} ({{\bf h}_{i}^{T}{\bf W}}),
\end{aligned}
\end{equation}

\begin{equation}
\begin{aligned}
\label{pplug}
&{P_{\rm plug}(y|{\bf{y}}_{<i}, {\bf{x}}, {\bf s}_1^{N}, {\bf t}_1^{N}; \bf {\theta})} \\
&=\left\{
\begin{array}{ll}
0, &{\text{if}}\ y \not\in {\rm \bf t}_1^{N} \\
\max{(0, \cos{(\frac{{\bf w}_y}{|{\bf w}_y|}, \frac{{\bf h}_i}{|{\bf h}_i|})})}, &{\text{if}}\ y \in {\rm \bf t}_1^{N}\\
\end{array} \right.
\end{aligned}
\end{equation}

\noindent where ${\bf h}_{i} \in \mathbb{R}^{d \times 1} $ is the hidden state of the $i$-th step from the last decoder layer, ${\rm W} \in \mathbb{R}^{d \times |\mathcal{V}|} $ is the embedding matrix, and ${\bf w}_{y}\in\mathbb{R}^{d \times 1}$ is the word embedding of token $y$. $P_{\rm plug}$ encourages the similarity between ${\bf h}_{i}$ and ${\bf w}_{y}$ for tokens inside the constraints.

Such formula has the problem of keeping the integrity of long constraints. It is possible that the cosine similarity between ${\bf h}_i$ and a word embedding from a wrong position is too high, causing the wrong token to appear in the $i$-th position. However, for long constraints, we have to ensure that all constraint tokens appear in the correct positions. To address this issue, we propose the integrity loss:

\begin{equation}
\label{ctrloss}
L_{\rm int} = -\sum_{y \in {\rm \bf{t}}_1^{N}} \log \frac{e^{\cos{(\frac{{\bf w}_y}{|{\bf w}_y|}, \frac{{\bf h}_i}{|{\bf h}_i|})}}}{\sum_{j=i-C}^{i+C}e^{\cos{(\frac{{\bf w}_y}{|{\bf w}_y|}, \frac{{\bf h}_j}{|{\bf h}_j|})}}}
\end{equation}

\noindent where $C$ is the window size. For each target token $y$ in the constraints, we use $C$ hidden states from the history and $C$ hidden states from the future as negative examples, our purpose is to prevent $y$ appears earlier or later in the translation. Finally, the training objective for VecConstNMT is: $L_{\rm origVecConstNMT} + \lambda L_{\rm int}$. The hyperparameter $\lambda$ is used to balance the original VecConstNMT loss and the integrity loss.

To further ensure the integrity of long constraints, we also propose gated decoding algorithm (GDA) for inference without sacrificing decoding speed. GDA tracks the decoding progress of each constraint and optimizes translation probability by a gating mechanism. The algorithm is presented in appendix \ref{appendix-gda} due to space limit.

\paragraph{Integration into The Template-based LCNMT} The template-based LCNMT \citep{wang2022template} uses the templates to simplify a sentence by disentangling different parts with different special tags. 
Formally, given a sentence pair and its $N$ lexical constraints, the template format is:

\begin{equation}
\begin{aligned}
\label{template1}
\begin{array}{rl}
& {\bf e}={\rm X}_0{\rm C}_1{\rm X}_1\cdot\cdot\cdot{\rm C}_N{\rm X}_N,\\
& {\bf f}={\rm Y}_0{\rm C}_{i_1}{\rm Y}_1\cdot\cdot\cdot{\rm C}_{i_N}{\rm Y}_N,\\
\end{array} 
\end{aligned}
\end{equation}

\noindent where ${\rm C}_1, ..., {\rm C}_N$ denote the slots for the source side constraints in order, similarly for ${\rm C}_{i1}, ..., {\rm C}_{iN}$ in the target side. ${\rm C}_{n}$ and ${\rm C}_{in}$ do not necessarily constitute a phrase pair. There is alignment between ${\rm C}_1, ..., {\rm C}_N$ and ${\rm C}_{i1}, ..., {\rm C}_{iN}$ that manifests the position relations between the constraints in the sentence pair. The $N$ lexical constraints divide the sentence pair into $N + 1$ textual fragments in each side, denoted by the nonterminals of ${\rm X}_0,..., {\rm X}_N$ in the source side and ${\rm Y}_0,..., {\rm Y}_N$ in the target side. 

The template provides clear configuration of the sentence pair. Since it reserves the slots for the constraints, the template based LCNMT 
guarantees the generation of the integral long constraints in the translation result. By using the slots for the constraints, we directly feed them the disambiguated constraints outputted by Stage 1 in the template based LCNMT at Stage 2.

\section{Experiments}

We conduct experiments on German-to-English (De-En) and English-to-Chinese (En-Zh) lexically constrained translation tasks. Different to major works on LCNMT that only use gold constraints, our experiment focuses on more practical scenario that ambiguous constraints exist given the input source sentences.

\subsection{Datasets}

\paragraph{Training Set} For De-En, the training set is from the WMT2014 German-English translation task, which consits of 4.51M parallel sentence pairs. For En-Zh, we construct the parallel training set from the corpora of WMT2021 shared task on machine translation using terminologies. Following \citet{wang-etal-2021-termmind}, we perform data selection based on in-domain n-gram match, which selects sentence pairs from all corpora that are similar to the task's validation set. After excluding the sentence pairs unrelated to the in-domain data, we use the 4.53M sentence pairs left as the training set.

\paragraph{Evaluation Set} For De-En, our test set is provided by \citet{wang-etal-2022-integrating}, which contains 508 sentence pairs with human-annotated alignments. Since the test set have significant overlaps with the corresponding training data, we remove all training examples which are covered by the test set. In addition, we use fast-align to annotate the newstest 2013 as the validation set. For En-Zh, both the test set and the validation set are provided by WMT2021 shared task on machine translation using terminologies, which consist of 2100 and 971 parallel sentence pairs respectively.

\begin{table}[!t]
\caption{Number of sentence pairs in each dataset. `Constrained' denotes the scenario where the sentence pairs contain the constraints, `Amb. Constrained' denotes the scenario where the sentence pairs contain the ambiguous constraints. \label{tab_data}}
\small
\centering
\begin{tabular}{c||c|c|c}
\hline
 & All & Constrained &  Amb. Constrained \\
\hline
\multicolumn{4}{c}{De-En} \\
\hline
Training & 4516710 & 3155213 & 2006279 \\
Validation & 3000 & 2049 & 986 \\
Test  & 508 & 318 & 203  \\
\hline
\multicolumn{4}{c}{En-Zh} \\
\hline
Training & 4535401 & 4511738 & 4477926 \\
Validation & 971 & 473 & 370 \\
Test  & 2100 & 1191 & 976  \\
\hline
\end{tabular}
\end{table}

\subsection{Lexical Constraints}

\begin{table*}[!t]
\caption{Main Results on De-En and En-Zh test sets. \label{tab_main}}
\small
\centering
\begin{tabular}{c||c|c|c|c|c|c}
\hline
Method & SacreBLEU & Exact-Match & CSR & Window 2 & Window 3 & 1 - TERm\\
\hline
\multicolumn{7}{c}{De-En} \\
\hline
Vanilla & 31.3 & 21.83 & 9.58 & 2.36 & 2.56 & 47.56 \\
Random + Stage2 Vec. & 34.4 & 40.61 & 46.12 & 9.51 & 9.44 & 49.17 \\
Most-Fre. + Stage2 Vec. & 33.8 & 41.76 & 47.09 & 9.47 & 9.61 & 48.72 \\
Ambiguous Vec. & 33.9 & 52.49 & 74.71 & 11.51 & 11.47 & 45.67 \\
Random + Stage2 Tem. & 34.8 & 52.56 & 56.55 & 17.07 & 17.67 & 49.67 \\
Most-Fre. + Stage2 Tem. & 34.6 & 53.79 & 57.77 & 17.29 & 17.46 & 49.54\\
Ambiguous Code-Switch & 34.9 & 71.85 & 75.31 & 14.67 & 15.03 & 48.27 \\
TermMind & 35.4 & 75.09 & 79.69 & 15.64 & 16.28 & 48.54 \\
\hline
Stage1 + Stage2 Vec. & 34.8 & 76.13 & 81.63 & 15.67 & 15.92 & 49.91 \\
Stage1 + Stage2 Tem. & \textbf{36.5} & \textbf{81.66} & \textbf{83.01} & \textbf{25.41} & \textbf{25.74} & \textbf{50.91} \\
\hline
\multicolumn{7}{c}{En-Zh} \\
\hline
Vanilla & 29.6 & 66.17 & 70.12 & 20.89 & 21.23 & 37.96 \\
Random + Stage2 Vec. & 29.5 & 69.13 & 73.86 & 20.29 & 20.87 & 37.61 \\
Most-Fre. + Stage2 Vec.  & 29.7 & 73.67 & 77.73 & 20.77 & 21.51 & 38.08 \\
Ambiguous Vec. & 29.4 & 70.17 & 80.44 & 20.91 & 21.29 & 38.53 \\
Random + Stage2 Tem. & 29.9 & 74.61 & 80.13 & 21.38 & 22.13 & 38.01 \\
Most-Fre. + Stage2 Tem. & 30.1 & 75.14 & 80.98 & 21.69 & 22.27 & 38.09 \\
Ambiguous Code-Switch & 27.1 & 65.46 & 75.19 & 18.13 & 18.71 & 28.54 \\
TermMind & 27.3 & 69.00 & 78.24 & 18.92 & 19.43 & 31.93 \\
\hline
Stage1 + Stage2 Vec. & 29.0 & 77.48 & 84.51 & 21.21 & 21.78 & 37.42 \\
Stage1 + Stage2 Tem. & \textbf{30.5} & \textbf{87.19} & \textbf{91.52} & \textbf{23.89} & \textbf{24.78} & \textbf{40.46} \\
\hline
\end{tabular}
\end{table*}

There are usually two ways to build the lexical constraints. One way is the simulation method adopted in most LCNMT researches \citep{chen2021lexically,wang2022template,wang-etal-2022-integrating}. They simulate the lexical constraints by extracting parallel phrases from the parallel sentences in both training and testing sets, and randomly selecting some parallel phrase as the lexical constraints. Such simulation method is not practical since we do not have parallel sentences during testing. In practice, it is usual that some source phrases have multiple possible translations, and constitute the ambiguous constraints. So, we simulate this practical scenario by collecting all possible translations of a considered source phrase as the ambiguous constraints. We study such simulated constraints in De-En. 


The other way is the human labeling method. WMT 2021 shared task on machine translation using terminologies provides manual translations of the source terminologies as the constraints. In comparison to the simulation method that is based on automatic word alignment and phrase extraction, the human labeling method builds the lexical constraints with higher quality. We study such human labeled constraints in En-Zh. Since the size of the human labeled terminology translation dictionary is too small for En-Zh training, we use the same strategy as the simulation method to extract the constraints in the training set. Following \citet{wang-etal-2022-integrating}, the number of constraints for each sentence in the training set is up to 3. 

Both the simulated constraints (in De-En experiment) and the human labeled constraints (in En-Zh experiment) have the ambiguity phenomena as shown in Table \ref{tab_data}. It shows that the sentence pairs containing ambiguous constraints account for majority of the sentence pairs that have constraints, indicating the wide spread of the ambiguous constraint phenomena. We are the first to conduct comprehensive studies on the constraints built by the two ways.

\subsection{Baselines}

We compare the proposed framework with the following baseline methods:
\begin{itemize}
\item{\textbf{Vanilla} We directly train a Transformer model \citep{vaswani2017attention} to translate, which is an unconstrained baseline.}
\item{\textbf{Random + Stage2 Vec.} At Stage 1, we randomly select one constraint from the ambiguous constraints for each considered source lexicon. At Stage 2, we inject the constraints of Stage 1 into VecConstNMT \citep{wang-etal-2022-integrating}.
\item{\textbf{Most-Fre. + Stage2 Vec.}} At Stage 1, for each considered source lexicon, we select its most frequent constraint in the training set as the constraints for VecConstNMT at Stage 2.}
\item{\textbf{Ambiguous Vec.} We directly feed all constraints for each considered source lexicon into VecConstNMT. This baseline does not explicitly disambiguate the constraints.}
\item{\textbf{Random + Stage2 Tem.} It is similar to "Random + Stage2 Vec.". The difference is that we use the template-based LCNMT \citep{wang2022template} instead of VecConstNMT at Stage 2.}
\item{\textbf{Most-Fre + Stage2 Tem.} It is Similar to "Most-Fre + Stage2 Vec.". The difference is the template-based LCNMT instead of VecConstNMT at Stage 2.}
\item{\textbf{Ambiguous Code-Switch} Similar to \citet{song-etal-2019-code}, we use the synthetic code-switching corpus to train the LCNMT model, the difference is that we use all constraints seperated by [SEP] to replace the corresponding source lexicon. }
\item{\textbf{TermMind} We use the data augmentation approach of TermMind, which is the winning system of WMT2021 machine translation using terminologies task \citep{wang-etal-2021-termmind}. It fuses ambiguous constraints into source sentences by special tags and masks source lexicon to strengthen the learning of constraints. }
\end{itemize}

\begin{table}[!t]
\caption{Results on Ambiguous Constraint Test Sets of De-En and En-Zh. \label{tab_many_simple}}
\small
\centering
\begin{tabular}{c||c|c}
\hline
Method & SacreBLEU & Exact-Match \\
\hline
\multicolumn{3}{c}{De-En} \\
\hline
Vanilla & 29.7 & 7.96 \\
Random + Stage2 Vec. & 31.4 & 16.81   \\
Most-Fre. + Stage2 Vec.  & 30.5 & 19.47   \\
Ambiguous Vec. & 30.7 & 42.36  \\
Random + Stage2 Tem. & 31.6 & 26.11   \\
Most-Fre. + Stage2 Tem. & 31.3 & 27.88   \\
Ambiguous Code-Switch & 32.6 & 56.46  \\
TermMind & 33.1 & 54.42  \\
\hline
Stage1 + Stage2 Vec. & 33.1 & 58.23  \\
Stage1 + Stage2 Tem. & \textbf{35.1} & \textbf{71.23}  \\
\hline
\multicolumn{3}{c}{En-Zh} \\
\hline
Vanilla & 29.6 & 65.40   \\
Random + Stage2 Vec. & 29.8 & 67.27   \\
Most-Fre. + Stage2 Vec.  & 29.7 & 71.58   \\
Ambiguous Vec. & 30.1 & 70.92   \\
Random + Stage2 Tem. & 30.1 & 75.43  \\
Most-Fre. + Stage2 Tem. & 30.4 & 76.49   \\
Ambiguous Code-Switch & 25.6 & 65.43   \\
TermMind & 25.9 & 65.56   \\
\hline
Stage1 + Stage2 Vec. & 29.2 & 76.23   \\
Stage1 + Stage2 Tem. & \textbf{31.4} & \textbf{84.59}   \\
\hline
\end{tabular}
\vspace{-1.5em}
\end{table}

\subsection{Evaluation metrics}
The evaluation includes constraint level and sentence level metrics. In the constraint level, we use metrics such as exact-match accuracy, which measures the appearance rate of the whole constraints in the translation results. In the sentence level, we use case-sensitive SacreBLEU \citep{post-2018-call}. Details of other metrics, including window overlap accuracy, terminology-biased translation edit rate (TERm), and CSR can be found in appendix \ref{appendix-metrics}.

\subsection{Results}

Table \ref{tab_main} presents the performances on the test sets of De-En and En-Zh. In each language pair, the top part lists the baseline performances, and the bottom part lists the performances of our two stage approach Stage1 + Stage2 Vec./Tem. It shows that our approach consistently outperforms baselines in both language pairs, especially leads a wide margin in constraint level evaluations. At the same time, our approach maintains or achieves higher sentence level SacreBLEU. Regarding two important constraint level metrics of exact match and CSR, which reflect the hard and soft accuracy of the constraints appeared in the translation result, our approach generally outperforms the strong baselines, including the strong data augmentation approach TermMind. The improvements are averagely nine points in exact match and averagely seven points in CSR. This indicates that our constraint disambiguation is effective that more accurate constraints are generated in the translation compared to the baselines or existing approaches, leading to significantly better user experience since the constraints usually carry key information.

The effect of the constraint disambiguation at Stage 1 is shown in the comparison between our approach and Random+Stage2 Vec./Tmp. or Most-Fre.+Stage2 Vec./Tmp., which randomly select the constraint or select the most frequent constraint at Stage 1, respectively. No matter which one we use from VecConstNMT or the template based LCNMT at Stage 2, our constraint disambiguation at Stage 1 is consistently better than the two baselines. Furthermore, our two stage approach with explicit constraint disambiguation at Stage 1 also performs significantly better than the baselines of conducting implicit disambiguation, i.e., Ambiguous Vec., Ambiguous Code-Switch, and TermMind. They just train the sequence-to-sequence model to implicitly select the appropriate constraints from all possible constraints. Regarding the comparison between VecConstNMT and the template based LCNMT at Stage 2, the template based one performs significantly better under the premise of the same Stage 1. Besides the constraint level evaluation, our two stage approach achieves better SacreBLEU on De-En and En-Zh than all data augmentation based approaches, including Ambiguous Code-Switch and TermMind.

\paragraph{On Ambiguous Constraint Test Sets}

As shown in Table \ref{tab_data}, not all constraints are ambiguous. To strictly investigate the effectiveness of our constraint disambiguation approach, we delete the sentence pairs that do not contain ambiguous constraints in the test sets. Table \ref{tab_many_simple} shows SacreBLEU and Exact Match on these new test sets. Full scores are presented in table \ref{tab_many} in the appendix. It exhibits the same trend to Table \ref{tab_main} with clear advantage of our approach over various baselines, especially in constraint level Exact Match. Our two stage approach is effective in producing correct constraints, performing much better than implicit disambiguation approaches of Ambiguous Vec./Code-Switch and TermMind.

\begin{table}[!t]
\caption{Comparison between our approach and WMT2021 Machine Translation Using Terminologies Shared Task participants. \label{tab_wmt21}}
\small
\centering
\begin{tabular}{c||c}
\hline
 System & Exact-Match \\
\hline
TermMind-sys2 & 85.6 \\
TermMind & 66.8 \\
LinguaCustodia-Sys1 & 82.9\\
LinguaCustodia-Sys2 & 82.9\\
LinguaCustodia-Sys1-v2 & 82.8\\
LinguaCustodia-Sys1-v3 & 82.8\\
KEP & 64.5  \\
\hline
Stage1 + Stage2 Tem. & \textbf{87.2} \\
\hline
\end{tabular}
\end{table}

\paragraph{Comparison to WMT 2021 Shared Task Participants}
We also compare our approach with the systems submitted to WMT 2021 shared task on machine translation using
terminologies in En-Zh. The systems are ranked according to Exact Match accuracy. Table ~\ref{tab_wmt21} shows that our Stage1 + Stage2 Tem. approach outperforms all participants. In addition, it is worth noting that TermMind-Sys2 uses techniques such as backtranslation, fine tuning on pseudo in-domain data and ensembling to enhance the performance of TermMind, while our approach does not add those techniques and only uses a subset of the training set, indicating the superiority of our approach on constraint disambiguation.

\section{Conclusion}
In this paper, we propose an effective two-stage framework for disambiguated lexically constrained neural machine translation (D-LCNMT). Our basic idea is to build a continuous representation space for constraint disambiguation at Stage 1, then inject the disambiguated constraints into the vectorized or template-based LCNMT models at Stage 2. Experiments show that our approach is significantly better than various representative systems across De-En and En-Zh translations, showing significant superiority in constraint disambiguation, which is wide spread and important in lexically constrained machine translation.

\section{Limitations}
In this paper, we does not specifically discuss morphological problems and polysemy problems, and does not develop special strategies for both problems such as \citet{pham-etal-2021-systran} and \citet{emelin-etal-2020-detecting}. Besides, the simulated lexical constraint dictionary, which is extracted from the parallel sentences of the training set based on automatic word alignment, may be different from the real lexical constraint dictionary provided by users.

\section{Ethics Statement}
D-LCNMT is designed as a machine translation system that can better serve the user pre-specified translation constraints. It can handle ambiguous constraints that are wide spread but neglected in major LCNMT researches. We believe that D-LCNMT would enhance user experience in machine translation services. In addition, the datasets used in our experiments are freely released data from WMT shared tasks.

\section{Acknowledgments}
We would like to thank the anonymous reviewers for the helpful comments. This work was supported by National Natural Science Foundation of China (Grant No. 62276179, 62261160648) and Alibaba Innovative Research Program.
\bibliography{anthology,custom}
\bibliographystyle{acl_natbib}

\appendix

\section{Appendix}
\label{sec:appendix}

\subsection{GDA}\label{appendix-gda}

Our algorithm uses next-tokens to record the translation of all constraints, which point the first ungenerated token for each constraint.
In step $i$ of decoding, we will judge whether the tokens generated in step $i-1$ are in next-tokens, if yes and the corresponding constraints are not fully generated either, we will update next-tokens and set the probabilities corresponding to the updated next-tokens in $P_{\rm plug}$ to 1. Unlike GBS, our method is not fully enforced, we use a gating mechanism to balance $P_{\rm model}$ and $P_{\rm plug}$. More importantly, our method does not hurt the decoding speed.
\begin{algorithm}[H]
\caption{GDA}\label{alg:gda}
\begin{algorithmic}

\STATE {\textbf{function}}$\ \mathbf{GDA} (\text{Max-Step}, \text{Beam-Size})$
\STATE \hspace{0.4cm}$ \text{beam} \gets \text{DECODER-INIT(\text{Beam-Size})}$
\STATE \hspace{0.4cm}$ \text{consts} \gets \text{[hyp.constraint for hyp in beam]}$
\STATE \hspace{0.4cm}$ \text{next-tokens} \gets \text{[c.next-token for c in consts]} $
\STATE \hspace{0.4cm}$ \textbf{for } \text{time step } i\ \text{in}\ \text{1,...,Max-Step} \ \textbf{do}  $
\STATE \hspace{0.8cm}$ \text{compute}\ P_{\rm model}$
\STATE \hspace{0.8cm}$ \text{compute}\ P_{\rm plug}$
\STATE \hspace{0.8cm}$ \textbf{if}\ \text{beam[-1] in next-tokens}\ \textbf{then}$
\STATE \hspace{1.2cm}$ \textbf{if}\ \text{is not finished(beam[-1])}\ \textbf{then}$
\STATE \hspace{1.6cm}$\text{next-tokens} \gets \text{update(next-tokens)}$
\STATE \hspace{1.6cm}$ P_{\rm plug}(y=\text{next-tokens}) \gets 1$
\STATE \hspace{0.8cm}$ \text{compute gate}\ g$
\STATE \hspace{0.8cm}$ \text{scores} \gets (1-g)P_{\rm model} + gP_{\rm plug}$
\STATE \hspace{0.8cm}$ \text{beam} \gets \text{BEST(scores, Beam-Size)}$

\end{algorithmic}
\label{alg1}
\end{algorithm}

\subsection{Metrics} \label{appendix-metrics}

In the constraint level, we adopt the metrics used in WMT2021 machine translation using terminologies task, including exact-match accuracy, window overlap accuracy and terminology-biased translation edit rate (TERm)\footnote{The evaluation scripts are available in: \url{https://github.com/mahfuzibnalam/terminology_evaluation}}. The exact-match accuracy measures the appearance rate of the whole constraints in the translation results. The window overlap accuracy measures the position accuracy of the constraints in the context window. TERm is an edit distance based metric for measuring the translation quality, especially tailored for the constraints. Details of these metrics can be referred to \citet{anastasopoulos2021evaluation}. In addition, following previous works \citep{chen2021lexically, wang-etal-2022-integrating}, we use the percentage of constraints that are successfully generated in translation as CSR, which differs from exact-match in that it does not require matching of whole constraint. In the sentence level, we report case-sensitive sacreBLEU \citep{post-2018-call} to evaluate the translation quality.

\subsection{Model Configuration}

Our models are implemented in Fairseq Library. At Stage 1, the hidden vector dimension of the disambiguation network is set 512. During training, we use Adam to optimize our models with $\beta_1=0.9$, $\beta_2=0.98$ and $\epsilon=10^{-9}$. The max learning rate is 0.0007 and warmup step is 4000. For each source lexicon, we randomly select 5 candidate translations as the negative samples.
At Stage 2, following previous works \citep{wang2022template,wang-etal-2022-integrating}, both VecConstNMT and the template-based LCNMT consist of 6 encoder layers and 6 decoder layers, and the hidden size is 512. Each multi-head attention module has 8 individual attention heads. During training, the learning strategy is the same to Stage 1. We set all dropout rates to 0.1. Besides, the hyperparameter $\lambda$, which is the weight of $L_{\rm int}$ in VecConstNMT, is set to 1 and the window size $C$ is fixed to 5. During inference, the beam size is set to 4. 

\begin{table*}[!t]
\caption{Results on Ambiguous Constraint Test Sets of De-En and En-Zh. \label{tab_many}}
\small
\centering
\begin{tabular}{c||c|c|c|c|c|c}
\hline
Method & SacreBLEU & Exact-Match & CSR & Window 2 & Window 3 & 1 - TERm\\
\hline
\multicolumn{7}{c}{De-En} \\
\hline
Vanilla & 29.7 & 7.96 & 16.84 & 4.43 & 4.42 & 44.23 \\
Random + Stage2 Vec. & 31.4 & 16.81 & 30.53 & 9.07 & 9.21 & 46.01 \\
Most-Fre. + Stage2 Vec.  & 30.5 & 19.47 & 30.88 & 11.51 & 10.86 & 44.08 \\
Ambiguous Vec. & 30.7 & 42.36 & 57.19 & 21.92 & 21.52 & 45.42 \\
Random + Stage2 Tem. & 31.6 & 26.11 & 37.19 & 12.36 & 13.48 & 46.92 \\
Most-Fre. + Stage2 Tem. & 31.3 & 27.88 & 38.95 & 12.45 & 12.59 & 46.45 \\
Ambiguous Code-Switch & 32.6 & 56.46 & 52.37 & 24.89 & 25.73 & 46.31 \\
TermMind & 33.1 & 54.42 & 58.25 & 26.99 & 27.92 & 47.25 \\
\hline
Stage1 + Stage2 Vec. & 33.1 & 58.23 & 69.12 & 28.37 & 28.02 & 47.59 \\
Stage1 + Stage2 Tem. & \textbf{35.1} & \textbf{71.23} & \textbf{75.44} & \textbf{32.45} & \textbf{32.71} & \textbf{48.95} \\
\hline
\multicolumn{7}{c}{En-Zh} \\
\hline
Vanilla & 29.6 & 65.40 & 72.54 & 37.48 & 38.00 & 42.53 \\
Random + Stage2 Vec. & 29.8 & 67.27 & 75.48 & 37.27 & 36.35 & 41.80 \\
Most-Fre. + Stage2 Vec.  & 29.7 & 71.58 & 78.67 & 35.81 & 37.36 & 42.55 \\
Ambiguous Vec. & 30.1 & 70.92 & 78.72 & 37.12 & 38.32 & 42.44 \\
Random + Stage2 Tem. & 30.1 & 75.43 & 80.36 & 38.53 & 39.41 & 41.63\\
Most-Fre. + Stage2 Tem. & 30.4 & 76.49 & 81.51 & 38.54 & 39.74 & 42.64 \\
Ambiguous Code-Switch & 25.6 & 65.43 & 73.27 & 30.67 & 31.88 & 30.12 \\
TermMind & 25.9 & 65.56 & 75.89 & 31.23 & 32.42 & 29.83 \\
\hline 
Stage1 + Stage2 Vec. & 29.2 & 76.23 & 83.44 & 36.39 & 37.51 & 40.98 \\
Stage1 + Stage2 Tem. & \textbf{31.4} & \textbf{84.59} & \textbf{89.65} & \textbf{42.01} & \textbf{42.88} & \textbf{44.84} \\
\hline
\end{tabular}
\end{table*}

\subsection{The effects of The Intergrity Loss and GDA for VecConstNMT}

\begin{table*}[!t]
\caption{Ablation study on VecConstNMT with different length constraints. \label{tab_ablation}}
\centering
\small
\begin{tabular}{c||c|c|c|c|c}
\hline
 Method & Metrics & Len.$=$1 & Len.$=$2 & Len.$=$3 & Len.$\geq$4  \\
\hline
\multirow{2}{*}{Orig. VecConstNMT} & CSR &95.86&98.79&93.06&100.00\\
 & Exact-Match &95.86&81.82&62.50&50.00 \\
\hline
\multirow{2}{*}{Ours} & CSR &97.63&97.58&95.83&100.00\\
 & Exact-Match &97.63&91.51&81.94&72.22 \\
\hline
\multirow{2}{*}{w/o GDA} & CSR &97.63&96.97&88.39&100.00\\
 & Exact-Match &97.63&88.48&76.39&66.67 \\
\hline
\multirow{2}{*}{w/o The Integrity Loss}& CSR  &95.27&96.97&93.06&100.00\\
 & Exact-Match &95.27&84.84&68.06&66.67\\
\hline
\end{tabular}
\end{table*}








Although VecConstNMT outperforms several strong baselines \citep{wang-etal-2022-integrating}, we found that the exact-match accuracy of the original VecConstNMT model decreases dramatically as constraint length increases. Table \ref{tab_ablation} shows the translation accuracy of the constraints with different lengths. We report CSR, which reflects the performance of the constraints at word level, and report exact-match which reflects the performance of the constraints at whole phrase level. It is obvious that exact-match decreases sharply with the increase of constraint length, but CSR remains stable. This phenomenon indicates that the original VecConstNMT can translate long constraints at word level, but it cannot make these constrained tokens appear consecutively in the correct order.

To address this problem, we propose the integrity loss during training and GDA during inference. To verify the effect of them on assurance of the long constraints, we experiment by using only the gold constraints in De-En translation. As shown in Table~\ref{tab_ablation}, 
the longer the constraint, the more significant improvement on exact-match can be achieved. In comparison to the original VecConstNMT, our model achieves 10-point improvements on constraints of length 2, and nearly 20-point improvements on constraints longer than 2. When we do not use the integrity loss, exact-match on length < 4 decreases more significantly than not using GDA, indicating the effectiveness of the integrity loss.



\subsection{The effect of Adding VDBA to Stage2 Vec.}

\begin{table*}[!t]
\caption{The results of adding VDBA to Stage2 Vec. \label{tab_vdba}}
\centering
\small
\begin{tabular}{c||c|c|c|c}
\hline
 & SacreBLEU & Exact-Match. & Window 2 & 1 - TERm \\
\hline
Random + Stage2 Vec. & 34.4 & 40.61 & 9.51 & 49.17  \\

+ VDBA & 33.1 & 52.08 & 15.80 & 46.32\\

Random + Stage2 Tem. & 34.8 & 52.56 & 17.07 &49.67\\

Most-fre. + Stage2 Vec. & 33.8 & 41.76 & 9.47 & 48.72  \\

+ VDBA & 32.5 & 52.32 & 15.18 & 45.93  \\

Most-fre. + Stage2 Tem. & 34.6 & 53.79 & 17.29 & 49.54\\

Stage 1 + Stage2 Vec. & 34.8 & 76.13 & 15.67 & 49.91  \\

+ VDBA & 32.4 & \textbf{82.75} & 13.56 & 44.56  \\

Stage 1 + Stage2 Tem. & \textbf{36.5} & 81.66 & \textbf{25.41} & \textbf{50.91} \\
\hline
\end{tabular}
\end{table*}

Following \citet{wang-etal-2022-integrating}, we add VDBA to the decoding of our Stage2 Vec. VDBA dynamically devotes part of the beam for constraint-related hypotheses at inference time, achieving high exact-match of the constraints. Table ~\ref{tab_vdba} shows the results of adding VDBA in De-En translation. Exact-match accuracy is significantly improved when VDBA is added.
Our Stage1 + Stage2 Vec. + VDBA achieves the best exact match accuracy, but the window overlap metric and 1-TERm drop by 2.1 and 5.3 compared to Stage1 + Stage2 Vec., respectively. In addition, the introduction of VDBA seriously harms SacreBLEU and slows down the decoding speed. It shows that although adding VDBA improves exact-match, it is harmful to sentence-level performance or constraint level context performances.
\subsection{Disambiguation Accuracy}

\begin{table}[!t]
\caption{The disambiguation accuracy of different methods. \label{tab_disacc}}
\centering
\small
\begin{tabular}{c||c|c|c}
\hline
 & Random & Most-Fre. & Stage1 \\
\hline
\multicolumn{4}{c}{All} \\
\hline
De-En & 53.4 & 54.5 & 81.3 \\
En-Zh & 46.4 & 56.9 & 79.2 \\
\hline
\multicolumn{4}{c}{Ambiguous} \\
\hline
De-En & 27.1 & 28.8 & 71.3 \\
En-Zh & 36.8 & 50.3 & 78.3 \\
\hline
\end{tabular}
\end{table}

We test the disambiguation accuracy on De-En and En-Zh language pairs. The results is shown in Tab.~\ref{tab_disacc}, where "All" presents all constraints in the orginal test sets and "Ambiguous" presents the constraints that each source terminology has muliple translation candidates. In Tab.~\ref{tab_disacc}, "Random" and "Most-Fre." denote selecting target translations at random and selecting translations with the highest frequency, respectively. We didn't report the disambiguation accuracy of data augmentation based methods, which cannot yield explicit disambiguation results to compute the accuracy. The experiment results demonstrate that our method outperform all baselines by a significant margin, especially on ambiguous constraint test sets.

\end{document}